# CancerKG.ORG–A Web-scale, Interactive, Verifiable Knowledge Graph-LLM Hybrid for Assisting with Optimal Cancer Treatment and Care

Michael Gubanov †
Florida State University
Tallahassee, FL, USA

Anna Pyayt
University of South Florida
Tampa, FL, USA

Aleksandra Karolak
Moffit Cancer Center and
Research Institute
Tampa, FL, USA

**Abstract**

Here, we describe one of the first Web-scale hybrid *Knowledge Graph (KG)-Large Language Model (LLM)*, populated with the latest peer-reviewed medical knowledge on colorectal Cancer. It is currently being evaluated to assist with both medical research and clinical information retrieval tasks at Moffitt Cancer Center, which is one of the top Cancer centers in the U.S. and in the world. Our hybrid is remarkable as it serves the user needs better than just an LLM, KG or a search-engine in isolation. LLMs as is are known to exhibit hallucinations and catastrophic forgetting as well as are trained on outdated corpora. The state of the art KGs, such as PrimeKG, cBioPortal, ChEMBL, NCBI, and other require manual curation, hence are quickly getting stale. CancerKG is unsupervised and is capable of automatically ingesting and organizing the latest medical findings. To alleviate the LLMs shortcomings, the verified KG serves as a Retrieval Augmented Generation (RAG) guardrail. CancerKG exhibits 5 different advanced user interfaces, each tailored to serve different data modalities better and more convenient for the user.

## 1 INTRODUCTION

Published peer-reviewed medical knowledge and practices double every few months [29]. This complicates quick access to it and hinders awareness of the latest best practices for all interested parties. Patients, their families, and medical professionals [38, 42–45]– all are forced into time-consuming Google/PubMed/QxMD/other search, followed by reading and filtering out multiple Web-pages, publications, etc, which is prohibitively slow[19, 35, 36, 39, 40, 62, 68, 70]. Our innovation is a hybrid Knowledge Graph (KG)-LLM that provides quick access to the latest personalized best practices and other latest medical findings, found in the latest peer-reviewed publications. It is a RAG-based system [15] comprised of an LLM (a choice of Meta Llama 2 [18], Google FLAN T5 [27], GPT-2 [71] or GPT-4 [61]) moderated by our trustworthy Knowledge Graph (KG). This hybrid marries the strengths of LLMs with verifiability and multi-modal content compatibility of our novel KG. Traditional KGs, Deep-learning models, or LLMs cannot be used to reliably retrieve and organize complex knowledge from thousands of publications, without significant human supervision to ensure correctness. LLMs require almost no supervision, but still suffer from other major AI-related limitations, such as "hallucinations" [2] and "catastrophic forgetting" [3], often leading to "forgetting" important information or inventing fake facts. Furthermore, most of them are trained on outdated data (e.g. cut-off date of September 2021 for GPT-4 [4]), and are very expensive to retrain.

Current socially maintained generic KGs, such as YAGO [65] or DBPedia [22]; medical ontologies and databases, such as NCBI, Viral [10] or PrimeKG[26]; Cancer databases, such as cBioPortal [25] or ChEMBL [33] are all manually curated, hence quickly become stale and have limited coverage. Other manually curated popular resources such as CDC.gov and WebMD.com are updated more frequently, but are very shallow, since the highly educated personnel can only afford to cover only the most dominating topics due to high cost.

This makes both traditional KGs and LLMs as is unsuitable for solving the problem. Our RAG-based hybrid scales to thousands of data sources, "understands" multi-modal knowledge, does not hallucinate, and does not require massive supervision. It learns from the latest peer-reviewed publications from PubMed.com and exhibits both broad topical coverage within the domain, as well as topical dept. It has several interactive interfaces – browsing, search, and natural language. This novel solution helps users access the latest relevant knowledge that is actionable for patient care. It currently undergoes evaluation and is expected to be reduced into medical research practice on colorectal Cancer patients. We take colorectal Cancer as a model domain, but without making our architecture depend on it, so the overall approach remains truly "on demand" – i.e. applicable to other scientific areas [17, 20, 24, 37, 46–48, 50, 54–59, 64, 67, 73].

†Corresponding author, gubanov@cs.fsu.edu

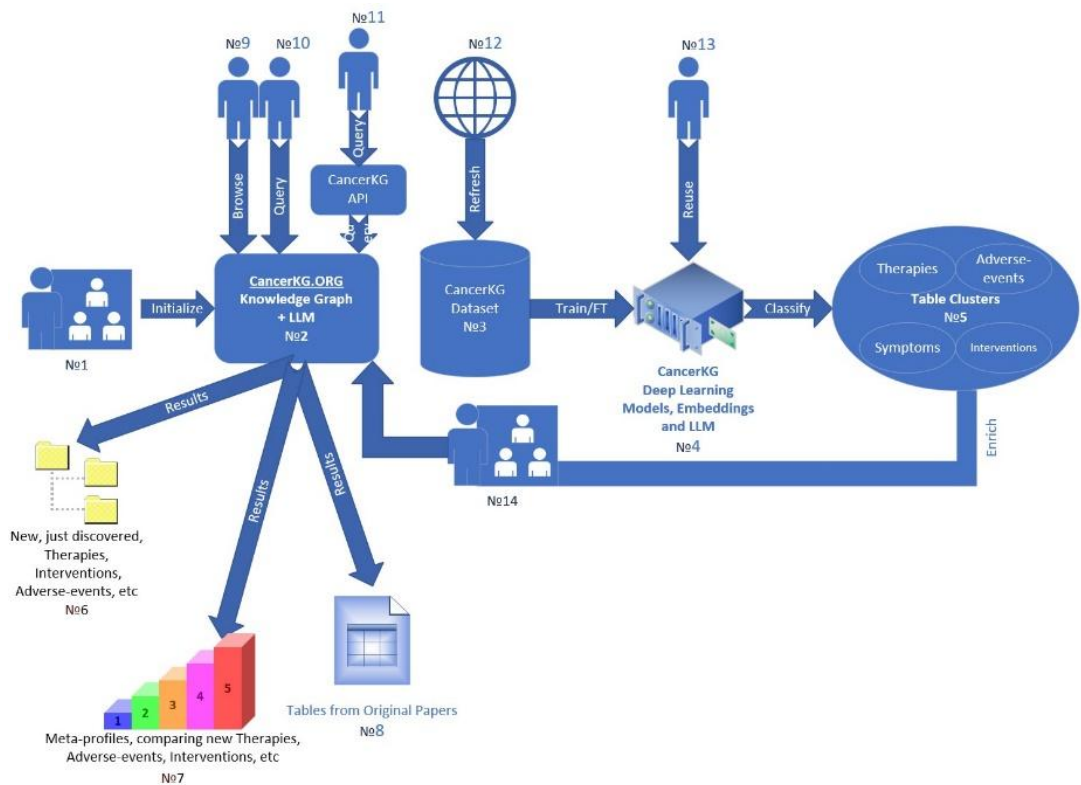

**Figure 1: CancerKG.ORG Architecture.**

We start by detailing the CancerKG architecture. Next, we describe some of our Deep-Learning models and the process that we designed to train them at scale. These models extract and organize knowledge from our datasets into our KG. Finally, we discuss how we fine-tuned our LLMs and how it interfaces with our KG to provide the hybrid balanced solution. We finish by reviewing related work and some similar systems currently used in practice in the Cancer Centers worldwide.

## 2 ARCHITECTURE

After researching the state of the art Knowledge Graphs [2, 22, 25, 26, 30, 32, 33, 65, 69, 75], LLMs [18, 27, 61, 71] as well as investigating the needs of colorectal Cancer patients, oncologists, and data scientists through conducting interviews during our NSF I-Corps customer discovery process [14], we have designed and tested the current architecture of CancerKG.ORG. It is similar to [76] with differences in it using Large Language Models (LMM), different Deep Learning models, Meta-profiles, and datasets. It is illustrated in Figure 1. №1 in the Figure represents a Data Scientist, who manually initializes a very small vetted KG with 10-20 nodes and interconnecting edges (depending on the domain), which will serve as a seed of our KG. №2 corresponds to our KG, stored in a scalable triplestore, such as Amazon Neptune [23], Eclipse RDF4J [7] or sharded MongoDB storage [12]. This KG is interactive and can be browsed (see Figure 2) or queried via publication (see Section 3) or table structural search-engines (see Figure 3) or API. №3 depicts our CancerKG dataset. It is parsed, post-processed, and restructured before storage in a semi-structured format (i.e. JSON), convenient for training Machine/Deep-learning models, Embeddings and fine-tunning LLMs. №4 represents a high-performance NVidia GPU cluster, responsible for training, classification, clustering, and LLM fine-tuning and question-answering (QA) workloads. It is configured with Apache Spark MLLib [21], TensorFlow [16], and LLM such as Meta LLama 2 [18], Google FLAN T5 [27], GPT-2 [71] and GPT-4 [61]. №5 in Figure 1 shows the topical table clusters, extracted and formed from the dataset. №6 illustrates a hierarchical KG fragment, learned from these clusters – in this case for (colorectal) Cancer - new therapies, adverse-events, symptoms, etc. №7 corresponds to the multi-layered 3D Meta-profiles, generated from these clusters. Meta-profile [41, 63] is a concise and convenient visualization/browsing interface that we proposed for accessing knowledge in large topical table clusters (see Figure 5). №8 corresponds to the tables from the original corpus. №9, 10 represent users, who browse, query, and ask CancerKG questions. №11, 13 are the CancerKG API users that use RPC or REST remote calls to do the same from their application or access our pre-trained Deep-learning models, Embeddings, or fine-tuned LLMs.

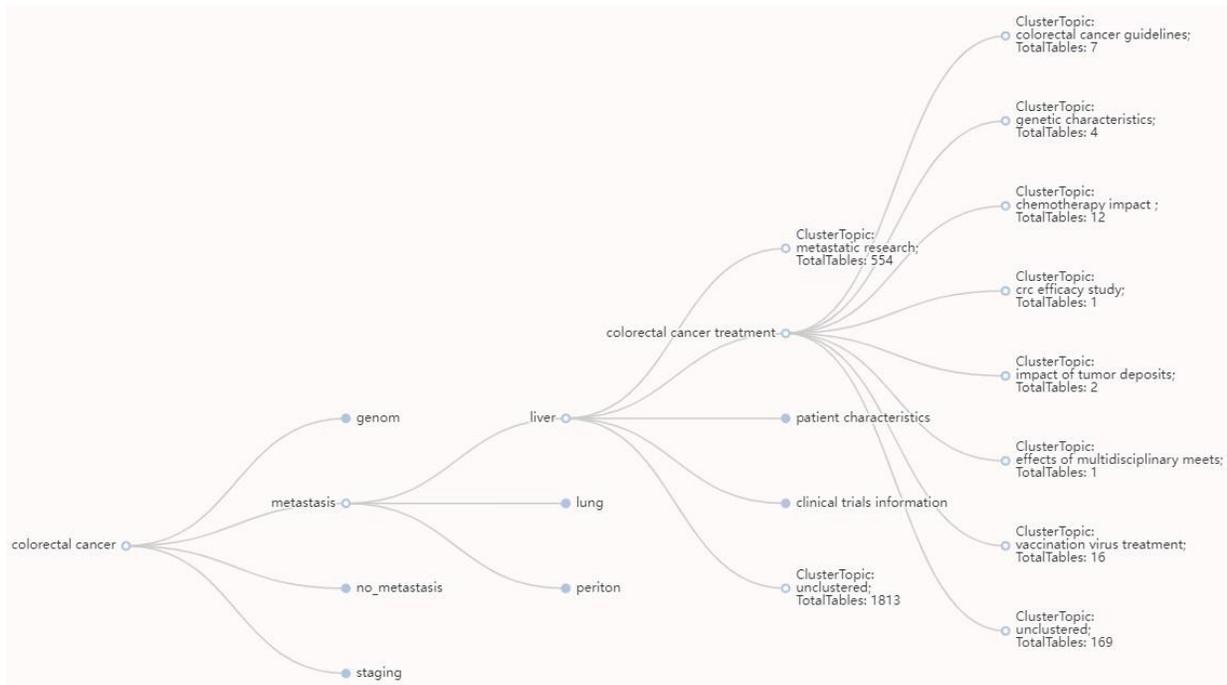

**Figure 2: A Small Fragment of the Interactive CancerKG Knowledge Graph.**

№12 depicts the World Wide Web (i.e. PubMed in our *case*) with the new vetted medical knowledge on colorectal Cancer published every day. №14 illustrates the KG enrichment process through fusion of new KG sub-trees or insertion of new nodes/edges.

Currently, CancerKG stores more than 2.5 million latest peer-reviewed publications on Cancer (including colorectal Cancer), parsed, decomposed in the KG, classified, and continuously updated with vetted knowledge from new peer-reviewed publications.

**Hardware, Libraries, Storage:** Training and validating of some of our models were done on a cluster of 4 machines, each having 4 Intel Xeon 2.4Ghz 40-core CPUs, from 192GB to 1TB of RAM, with 10TB disk space each, interconnected with a 1GB Ethernet. LLM fine-tunning was done using Amazon P5 AWS instances with NVidia latest GPUs. All software was written in the Python programming language. For implementing the RNN, GRU, and biL- STM models, we have used Keras, with Tensorflow framework as the backend. Our MongoDB [12] sharded cluster storing data and all trained Deep-learning models and embeddings takes approximately 965GB for its distributed dataset storage, with raw space consumption of more than 5TB.

## 3 INTERACTING WITH CANCERKG

CancerKG has several advanced user interfaces – an interactive KG, publication and table structural search engines, conversational interface in natural language, and 3D Meta-profile interface.

### 3.1 Interactive Knowledge Graph

Figure 2 illustrates the interactive KG that allows convenient inter- action with the hierarchical knowledge base learned by CancerKG. In Figure 2, the user clicks on the nodes and follows the unfolding path through *metastasis*, *liver*, *colorectal cancer treatmen*t nodes to the nested leaf-nodes having the topical clusters of tables connected to the corresponding KG leaf-nodes. After clicking on the leaf node, the user can choose "Show all tables" option from the contextual pop-up menu, which will display them in the bottom frame under the KG shown in Figure 3. That interface supports both drilling down deeper in the cluster using either the structural search or conversational interface. Alternatively, the user can choose "3D-meta profile" option in the same pop-up menu, which will generate a Meta-profile corresponding to the selected cluster. The Meta-profile generated for "Summaries and Case Studies" cluster is illustrated in Figure 5.

## 3.2 Search-engines

**Figure 3: CancerKG Structural Search over Tables based on the LLM Conversational Interface.**

We currently support 2 publication search-engines and one tabular search-engine, coupled with a conversational interface. During query processing we tokenize the query and perform stemming. Our ranking function features include the number of matches, proximity between the matched terms, relative importance of the matched field, term, etc. Each term (its NLP "root form") in the corpus has an associated Term Frequency-Inverse Document Frequency (TF-IDF) [52] weight in order to reward more important terms. For each matched term its TF-IDF is weighted in the ranking per document. *The first publication search-engine can be used to search separately over title, abstract, body text, and/or table captions, table data and metadata, figure captions and content*. It is more robust compared to a standard keyword-search over the entire publication (e.g. Pubmed.com, Google Scholar, etc) and allows more fine-grained filtering capabilities. The search fields are inclusive in the search results, meaning, if a user searches on a field there must be a document that matches at least one term in that field or it does not get passed on to the next stage regardless if there are matches over the other fields. The results are formatted with table captions first, the title and authors and the full abstract.

*The second publication search-engine* performs query processing differently – i.e. it matches the query terms to all fields used by the first search-engine above. It can be used whenever the user is unsure of where exactly the term may be. These search results are formatted with a brief excerpt of where it matches to the fields. The interface also allows the user to expand and collapse sections of the paper displayed in the search-results to get to the needed information quicker.

*The tabular search-engine* allows the user to search over a set of popular and important for clinical use table attributes present in the dataset. This is to our knowledge, one of the first structural search-engines over medical tables like that. Other solutions do not specifically separate tables, which leads to inability of querying their fields separately from the rest of the publication data or do not "understand" the intricacies of complex medical non-relational tables [53]. Generic relational [28] and semi-structured databases [12] can be used to load tables and use SQL to query them, but they do not "understand" intricacies of structure of such tables, hence fail to correctly support many challenging data harmonization tasks that are necessary to support correct and efficient structural search over such tables. They are usually not in 1st Normal Form [28], exhibit not only horizontal (HMD), but also vertical metadata (VMD) [53]. To query such tables efficiently many steps related to processing their non-standard structure with both HMD, VMD, and nesting have to be done correctly. Such steps include hierarchical vertical and horizontal schema matching, data transformation and unification, processing the nested tables inside the cells with their own metadata correctly (i.e. just unnesting would not help in this case), ranking search-results containing such tables by relevance – all of it is not addressed in the entirety in any of the solutions to our knowledge.

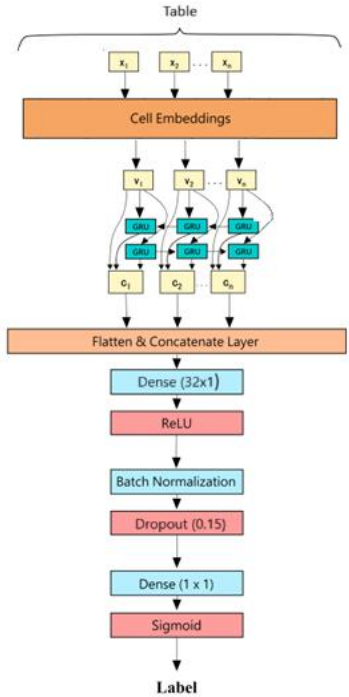

**Figure 4: Deep-learning Architecture for Topical Table Classification.**

Query processing over complex tables is a large part of our long-term goal to advance structural information retrieval for structured data at scale. These search results are a product of an advanced tandem of embedding-based schema matching (e.g. *Tumor Size, Effect Size, Size*) and advanced query processing. Figure 3 depicts a screenshot of search results for tables evaluating clinical outcomes with risk factors for colorectal Cancer. The user (e.g. oncologist) enters the natural language query in the conversational interface (its description is below) *"output all latest information available about risk factors and predictive models for metastatic colorectal cancer with tumor in lymph node, size 8.45"*. The query got parsed by our conversational query parser and got split into two queries. The first - a structural query that consists of the extracted attributes - *lymph node* and *tumor size* 8.45 and is executed via the table search engine. The second query is textual and is equivalent to the input, amended with the synonyms for the identified table fields to simplify query processing for the LLM. Both matching to the fields and synonym amendment is done using our custom embeddings that we trained on tables in the dataset. Figure 3 displays synonyms (both for the search terms and the attribute names) in the dark grey message box that we enabled for demonstration purposes. The screenshot is cut off at the bottom due to space constraints.

### 3.3 Conversational Interface

We fine-tuned several LLMs, such as LLama 2 [18], GPT-2 [71], GPT-4 [61], FLAN-T5 [27] on our corpus and offer a conversational interface in natural language to the user. The query is first being parsed by our conversational query parser, which identifies any table attributes and their values (if present) and automatically fills out the fields in the structural table search-engine. Second, the query is passed on further to one of the LLMs selected by the user, which generates a natural language response amending the tables (if any) generated by the search-engine. Figure 3 on the right, illustrates a user asking a question and the LLM reformulating the query for the search-engine that outputs the response (a table).

### 3.4 3D Meta-profile

A Meta-profile, informally, is a summary of metadata of a table cluster. Since, here our tables have both HMD and VMD, the metaprofile summarizes them in two separate dimensions. Figure 5 illustrates a Meta-profile generated by the user, who was browsing the Knowledge Graph, drilled down to the "Summaries and Case Studies" leaf node and selected an option "Create a 3D-meta Profile" from the pop-up menu. The Meta-profile is a 3D-bar graph that on X-axis has attribute labels of HMD and VMD of tables from the cluster, selected by the user. On Y-axis it has the TF/IDF [52] score corresponding to each HMD or VMD attribute. By clicking on the bars (blue corresponds to HMD, red - to VMD), the user can further drill down to the subset of tables from the cluster specifically having only the selected attributes corresponding to the selected bars. In other words, it can be thought of as a dynamic filter, creating new table sub-clusters based on the HMD and VMD choices made by the user. For example, if the user selects the "study design" blue bar in Figure 5, the system will generate a separate

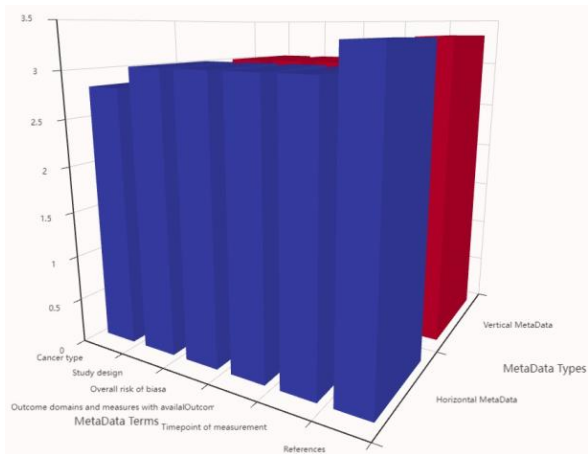

**Figure 5: A Meta-Profile generated by CancerKG.ORG for Summaries and Case Studies on colorectal Cancer.**

table sub-cluster, having only the tables from the original cluster having "study design" in their HMD. CancerKG will create such sub-clusters on the fly and amend the KG by attaching it to the original cluster.

## 4 TABLE TOPICAL CLASSIFICATION

### 4.1 GRU Model

Figure 4 depicts the architecture of a GRU architecture that we used for topical table classification, consisting of three main stages. In the first stage, a table, $\{x_1, x_2, ..., x_n\}$, where $x_i$ is the $i^{th}$ term is pre-processed to create cell-wise representations. It includes data cleaning along with the replacement of numbers and ranges in data with placeholders such as *NUM*, *RANGE*, etc as is described above. The pre-processed feature vectors are then used to fine-tune BioBERT embeddings [31] on the whole corpus. This sequence is passed through a GRU layer and the result is concatenated with the original embeddings to create our enriched contextualized vectors, $\{c_1, c_2, ..., c_n\}$. The final stage of the model passes them through a dense layer of 32 units, a batch normalization layer, a dropout layer and a dense binary classifier similar to [76].

### 4.2 Pre-processing

We have used 100,000 dimensional feature space, i.e. 100K English terms in our vocabulary that we have selected by taking all terms from our datasets, sorting by frequency and cutting off the noise words and spam [73]. Increasing the dimensionality further led to significantly slower training time, which would prevent or make the experiments much more difficult.

To streamline the processing of numerical data handled by the model, we have created several regular expressions that encode all numerical data falling in similar forms under its relevant category. The substitution is described in more detail in [53].

### 4.3 Training and Evaluation

We composed the training sets from Web-scale datasets such as WDC [60] and CancerKG respectively [74]. We evaluated our models and observed approximately 95% F-measure, when validated with 10-fold cross-validation, for Deep-learning Bi-GRU- based models with slight differences depending on whether the classified metadata is horizontal or vertical, as well as its position.

We composed the training sets for the topics corresponding to the leaf nodes in the colorectal Cancer KG by, first, asking a clinical Data Scientist to select a representative (i.e. centroid) table for each topic that we further use a seed to train our GRU binary classification model and create a cluster for each topic. Second, we created a composite embedding vector corresponding to each topical centroid table. Each table vector consists of 3 concatenated components - $V_{HMD}$ for HMD, $V_{VMD}$ for VMD, $V_D$ for D. Each of them is calculated as a summation of our embedding vectors (we fine-tuned BioBERT on our dataset [31]) corresponding to each term located in one the table sections. The final embedding vector for a table is a concatenation of 3 vectors – $V_T = V_{HMD} \oplus V_{VMD} \oplus V_D$. Third, we take a centroid vector and select only the tables in the dataset within a 18 degrees from it (determined empirically). Fourth, we train our GRU model as a binary topic classifier on these tables (as positively labeled) amended with the same number of random tables from the dataset (as negatively labeled). Finally, we run all such topical binary classifiers through our dataset on colorectal Cancer and form the table clusters of varying size.

## 5 KNOWLEDGE GRAPH

### 5.1 Initialization

The structural hierarchy (i.e. nodes and edges) for the Knowledge Graph will be initialized with the help of a Data Scientist (№1 in Figure 1). On the highest level, the general characteristics of Cancer can be extracted from PrimeKG [26], vetted static ontologies or dictionaries on colorectal Cancer. Once initialized, the KG gets automatically updated from the vetted medical sources. This ensures reliability, freshness, and quality of our KG (i.e. №2 in Figure 1).

### 5.2 Enrichment and Fusion

Once the KG initialization is complete we fuse the extracted information into our Knowledge Graph during the enrichment process. We classify and extract the clusters on prominent topics in colorectal Cancer (e.g. №5 in Figure 1). This process is challenging since all topical clusters have different structure and significant concepts and terms can be referred to differently (e.g. *mCRC* and *metastatic colorectal Cancer*). Consequently, we trained a variety of advanced AI models with our new tabular embeddings to help perform accurate clustering [34, 57].

The graph is populated with nodes and edges and is stored in JSON format. The structure of the graph is hierarchical, so all child nodes have parent nodes. The user can search over the KG via the front-end interface that except matching nodes also highlights the path to the matching nodes. The user can then either browse the graph to explore the table clusters attached to the nodes; after selecting a cluster clicking *show tables* or *3D metaprofile* in the popup menu to query the tables in the cluster (see Figure 3) or generate a meta-profile corresponding to the cluster (see Figure 5).

Fusion of the extracted hierarchical knowledge into a segment or several segments of our KG requires taking into consideration multiple levels of abstraction. For example, "Symptoms" can be a node in a subtree "Clinical presentation" that could be, in turn, linked to the "colorectal Cancer" KG root node. Because of the different ways to categorize, the actual symptoms may overlap in different KG subtrees. After consulting with several medical experts it was decided to store all different ways to categorize the data without merging them, since each of the categorization methods can be useful for different tasks that oncologists, trainees, and data scientists perform. While general public might be interested in common and rare symptoms, medical specialists might analyze specific organ systems. For example, sorting by "rare symptoms" and "common symptoms" can overlap with the sets of symptoms sorted by "organ systems". In addition to that, even though "Neurological symptoms" are related to the nervous system in general, while "Cerebrovascular" is related to the brain and its blood vessels, they have significant overlap in symptoms. The first step of fusing the extracted hierarchical knowledge into the KG is matching the root node of the extracted subtree to the corresponding node(s) in the KG. This matching process is based on normalized NLP term matching, amended by the embedding-driven matching. The latter is especially important in context of new terms, unseen before, which is often the case with new therapies, adverse-events, etc. For example, assume we have extracted a subtree 2nd line Treatments → Regorafenib from the table's metadata. The root node Therapy may match to the KG node Therapy(ies) by normalized NLP term matching and then the leaves (Regorafenib) can be merged with the leaves of the matched node in the KG. However, if there is no corresponding KG node Therapy(ies) and there is no match to the KG leaves with existing therapies, the embedding vector corresponding to the new therapy (Regorafenib) extracted from metadata can be used to match it to the embeddings vectors of the existing therapies in the KG due to them being close to each other by distance. The node Therapy then can be added to the KG on the top of the Regorafenib node. If the extracted subtree has several layers of hierarchy, e.g. Side-effects → Pediatric side-effects → Severe pain, it has to be left separate from the existing side-effects in the KG, even if matched to them by having close embedding vectors. This is because, it is categorized as Pediatric side-effects, which is a separate category from regular side-effects, so both the new node Pediatric side-effects and its leaves have to be added to the KG, even if some of the side-effects overlap with the general side-effects, already present in the KG. Fusion of sub-trees, having several layers or insertion of new nodes matching with a low confidence score has to be evaluated by an expert (№14 in Figure 1); fusion of leaves with nodes matched with high confidence score may be left unsupervised. Over time, all categories of initial fusion mistakes identified by the expert will be learned by the fusion model to be automatically corrected, hence most of the fusion is expected to become unsupervised in long run.

## 6 RELATED WORK

[8] is an Information Retrieval (IR) system over publications at researchrabbit.ai. They are introducing a retrieval mechanism over papers that does not require the use of keyword-search. They dis- play a force directed graph of related, cited and referenced papers that a user can construct and use. They provide many features, such as being able to create your own custom graph of papers, curated collections to improve recommendations, personalized alerts, sharing and collaborating of papers and graphs, and among others the ability to discover author networks.

PrimeKG [26] is a free KG populated from certain classic, well- known medical ontologies such as NCBI, UniProt and ≈20 other legacy databases [25, 33]. Their graph has information on genes of interest, transcripts, protein identifiers function names and gene names and is manually curated. The latter limits its scalability, and hence its breadth, depth of coverage as well as freshness of medical knowledge.

Manually curated chemical databases such as ChEMBL [33], Zinc [11], Enamine [5] storing drug-like molecules, including the commercially available or hit molecules and their targets support advanced structural search for well-established drugs and molecules. Pharmaceutical knowledge bases - DrugBank [4], BindingDB [1], and Protein Data Bank (PDB) [6], the latter particularly for the structural biology support advanced search for well-established drugs and their interaction patterns. Both chemical and pharma- ceutical databases, however, are manually curated, which makes them unattractive compared to any solution, including ours, that is capable of ingesting/organizing the latest knowledge automatically.

The following systems, even though on COVID-19 are relevant as they similarly to us develop advanced query-processing or search- engines over scientific publications and their multi-modal content. A system by the Center for Artificial Intelligent Research, HKUST [72] is a free service and that utilizes NLP to support question answering along with the summarization to help discover relevant scientific literature on COVID-19. Their system is comprised of 3 modules. The pipeline begins with a user query sent through the first module - document retrieval, which does paraphrasing and search. Query paraphrasing converts the user query to several shorter and simpler analogous questions. The search-engine uses Lucene [13] to retrieve related publications. Then the snippet selector module finds the related evidence among the whole text by using the answer re-ranking, and term matching score. Finally, a "Multi-document abstractive summarizer" synthesizes the final answer from multiple retrieved snippets.

Another relevant system - [9] enables access to a COVID-19 Intelligent Insight portal of over 100,000 curated scientific publica- tions. Sinequa's search engine supports full-text search using NLP. The Search engine supports ranking by relevance and recognizes synonymy in their ranking function. The interface has 3 sections - one for the matched scientific results, one for showing more details on a selected result and the last one for filtering and sorting the result set. The system highlights important information throughout each result and tags them all by different classification labels. Sinequa's system is also provided for free.

COVIDScholar [3] is an information retrieval resource for COVID-19 and related scientific research, established by Matscholar's re-search effort. COVIDScholar also uses NLP. The query terms are matched against the title and abstract. COVIDScholar displays title, authors, abstract, and provides a link to the paper full-text at its original publisher with the list of related works. The system neither has a Knowledge Graph, LLM or an advanced structural search engine.

CancerKG organizes the documents and tables by topics into a Knowledge Graph (KG) and provides a wide variety of advanced interfaces - an interactive KG, several search-engines, including a structural search engine over tables, interactive 3D Meta-profiles, a fine-tuned, verifiable LLM interacting in natural language with the user. The user can either browse or search the KG, all sections of the original publication, title, abstract, table caption or query the tables. Except search, CancerKG supports structural query processing with embedding-based matching of query terms and queried attributes to the data or metadata sections of the tables. The search-results page provides a list of ranked scientific resources with access to each full-text of each section of the paper, full-text of the whole document, and a ranked list of tables with the most relevant results. The ranking function incorporates matching terms and synonyms, proximity, document, terms, publication trustworthiness, and the number of citations as well as others. The advanced search-engine over tables displays a brief section with the most relevant tables first that can be expanded to see more results.

## 7 CONCLUSION

Here we described CancerKG - the first, interactive, trustworthy, scalable Knowledge Graph/LLM RAG hybrid on colorectal Cancer. We take colorectal Cancer as a model, but without making our architecture depend on it, so the overall approach remains applicable to any other scientific and medical domains, given the models are retrained for that domain. The content is extracted and updated in unsupervised manner from PubMed.com that contains vetted, peer-reviewed medical publications. Hence is verifiable and contains the most up to date medical practices.

### Acknowledgements
This material is based upon work supported by NSF and Casey DeSantis Department of Health Florida Cancer Innovation Fund under Grant Nos. 2345794 and MOABG. We thank Nick Piraino, Kiran Muppana, Maitry Chauhan, Bhim Kandibedala for their contributions to earlier versions of this work..